\documentclass[runningheads,a4paper]{llncs}
\usepackage{amssymb}
\usepackage{wrapfig,lipsum,booktabs}
\usepackage{graphicx}
\usepackage{algorithm}
\usepackage{algorithmicx}
\usepackage{algpseudocode}
\usepackage{amsmath}
\usepackage{multicol}
\usepackage{multirow}
\usepackage{array}
\newcolumntype{C}[1]{>{\centering\let\newline\\\arraybackslash\hspace{0pt}}m{#1}}
\floatname{algorithm}{Algorithm}
\newcommand{\keywords}[1]{\par\addvspace\baselineskip
\noindent\keywordname\enspace\ignorespaces#1}

\begin{document}
\mainmatter  

\title{Large-scale Collaborative Imaging Genetics Studies of Risk Genetic Factors for Alzheimer's Disease Across Multiple Institutions}
\titlerunning{Large-scale Collaborative Imaging Genetics Studies}

%
%
\author{Qingyang Li$^1$, Tao Yang$^1$, Liang Zhan$^2$, Derrek Paul Hibar$^3$, Neda Jahanshad$^3$,
\\Yalin Wang$^1$, Jieping Ye$^4$, Paul M. Thompson$^3$, Jie Wang$^4$}
\authorrunning{Q. Li et al.}

\institute{
$^1$School of Computing, Informatics, and Decision Systems Engineering, Arizona State Univ.,Tempe, AZ; $^2$Dept. of Engineering \& Technology, Univ. of Wisconsin-Stout, Menomonie, WI; $^3$Imaging Genetics Center, Institute for Neuroimaging and Informatics, Univ.of Southern California, Marina del Rey, CA; $^4$Dept. of Computational Medicine and Bioinformatics, Univ. of Michigan, Ann Arbor, MI}

\toctitle{Large-scale Collaborative Imaging Genetics Studies}
\tocauthor{Authors' Instructions}
\maketitle
\begin{abstract}
Genome-wide association studies (GWAS) offer new opportunities to identify genetic risk factors for Alzheimer's disease (AD). Recently, collaborative efforts across different institutions emerged that enhance the power of many existing techniques on individual institution data. However, a major barrier to collaborative studies of GWAS is that many institutions need to preserve individual data privacy. To address this challenge, we propose a novel distributed framework, termed Local Query Model (LQM) to detect risk SNPs for AD across multiple research institutions. To accelerate the learning process, we propose a Distributed Enhanced Dual Polytope Projection (D-EDPP) screening rule to identify irrelevant features and remove them from the optimization. To the best of our knowledge, this is the first successful run of the computationally intensive model selection procedure to learn a consistent model across different institutions without compromising their privacy while ranking the SNPs that may collectively affect AD. Empirical studies are conducted on 809 subjects with 5.9 million SNP features which are distributed across three individual institutions. D-EDPP achieved a 66-fold speed-up by effectively identifying irrelevant features.

\end{abstract}

\section{Introduction}
Alzheimer's Disease (AD) is a severe and growing worldwide health problem.  Many techniques have been developed to investigate AD, such as magnetic resonance imaging (MRI) and genome-wide association studies (GWAS), which are powerful neuroimaging modalities to identify preclinical and clinical AD patients. GWAS \cite{harold2009genome} are achieving great success in finding single nucleotide polymorphisms (SNPs) associated with AD. For example, APOE is a highly prevalent AD risk gene, and each copy of the adverse variant is associated with a 3-fold increase in AD risk. The Alzheimer's Disease Neuroimaging Initiative (ADNI) collects neuroimaging and genomic data from elderly individuals across North America. However, processing and integrating genetic data across different institutions is challenging. Each institution may wish to collaborate with others, but often legal or ethical regulations restrict access to individual data, to avoid compromising data privacy.

Some studies, such as ADNI, share genomic data publicly under certain conditions, but more commonly, each participating institution may be required to keep their genomic data private, so collecting all data together may not be feasible. To deal with this challenge, we proposed a novel distributed framework, termed Local Query Model (LQM), to perform the Lasso regression analysis in a distributed manner, learning genetic risk factors without accessing others' data. However, applying LQM for model selection---such as stability selection---can be very time consuming on a large-scale data set. To speed up the learning process, we proposed a family of distributed safe screening rules (D-SAFE and D-EDPP) to identify irrelevant features and remove them from the optimization without sacrificing accuracy. Next, LQM is employed on the reduced data matrix to train the model so that each institution obtains top risk genes for AD by stability selection on the learnt model without revealing its own data set. We evaluate our method on the ADNI GWAS data, which contains 809 subjects with 5,906,152 SNP features, involving a 80 GB data matrix with approximate 42 billion nonzero elements, distributed across three research institutions. Empirical evaluations demonstrate a speedup of 66-fold gained by D-EDPP, compared to LQM without D-EDPP. Stability selection results show that proposed framework ranked \textit{APOE} as the first risk SNPs among all features.

\section{Data processing}

\subsection{ADNI GWAS data}
The ADNI GWAS data contains genotype information for each of the 809 ADNI participants, which consist of 128 patients with AD, 415 with mild cognitive impairment (MCI), 266 cognitively normal (CN). SNPs at approximately 5.9 million specific loci are recorded for each participant. We encode SNPs with the coding scheme in \cite{sasieni1997genotypes} and apply Minor Allele Frequency (MAF) $<0.05$ and Genotype Quality (GQ) $<45$ as two quality control criteria to filter high quality SNPs features, the details refer to \cite{yang2015detecting}.

\subsection{Data partition}

Lasso \cite{tibshirani1996regression} is a widely-used regression technique to find sparse representations of data, or predictive models. Standard Lasso takes the form of
\begin{equation}
\min_{x} \frac{1}{2}||Ax-y||^2_2+\lambda||x||_1: x\in \mathbb{R}^p,
\label{eq:1}
\end{equation}
where $A$ is genomic data sets distributed across different institutions, $y$ is the response vector (e.g., hippocampus volume or disease status), $x$ is sparse representa-tion---shared across all institutions and $\lambda$ is a positive regularization parameter.

\begin{figure}[t]
\centering
\includegraphics[height=4cm]{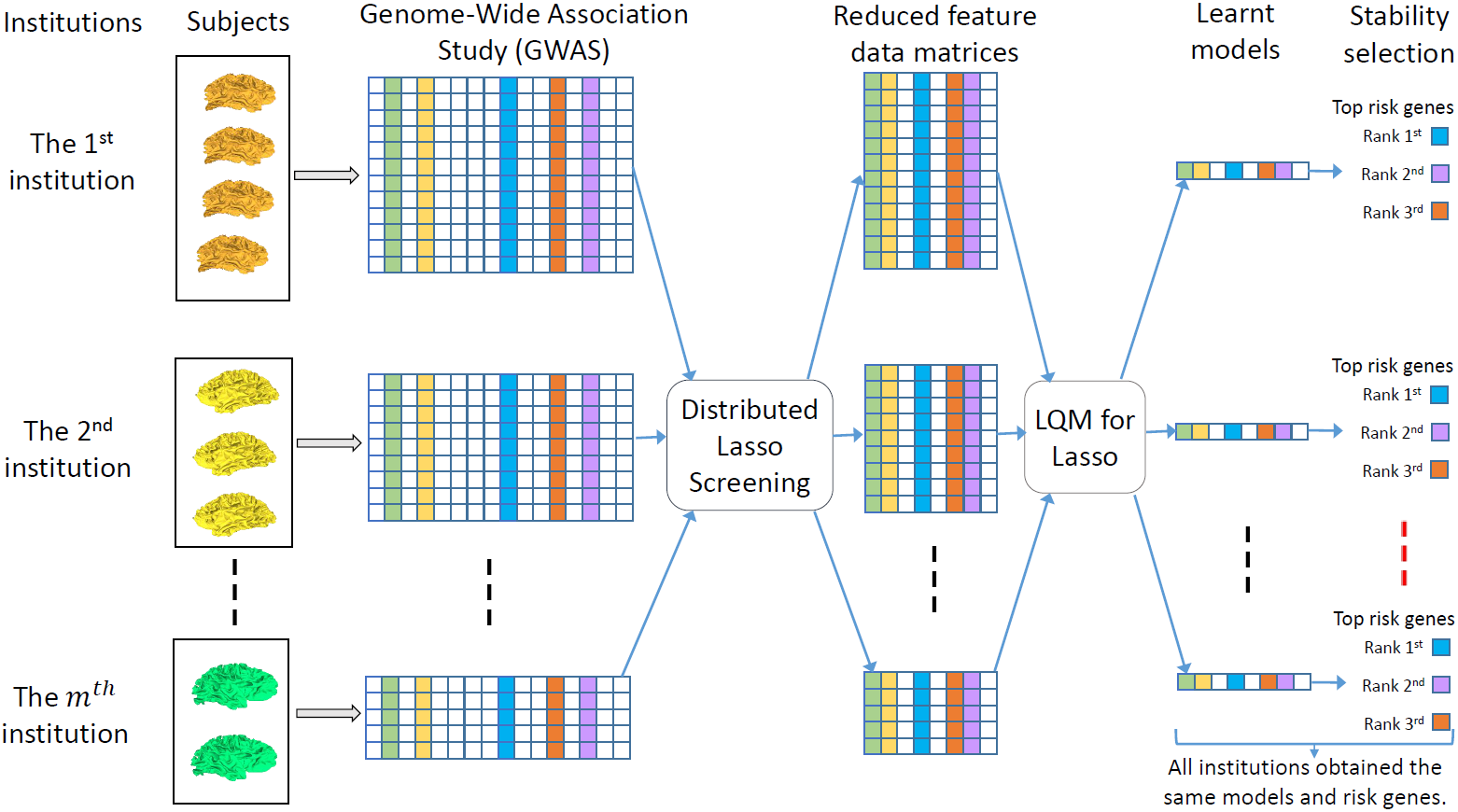}
\caption{The streamline of our proposed framework.}
\label{fig:1}
\end{figure}

Suppose that we have $m$ participating institutions. For the $i$th institution, we denote its data set by $(A_i,y_i)$, where $A_i\in \mathbb{R}^{n_i\times p}$, $n_i$ is the number of subjects in this institution, $p$ is the number of features, and $y_i\in \mathbb{R}^{n_i}$ is the corresponding response vector, and $n=\sum_i^m n_i$. We assume $p$ is the same across all $m$ institutions. Our goal is to apply Lasso to rank risk SNPs of AD based on the distributed data sets $(A_i,y_i)$, $i=1,2,...,m$.

\section{Methods}

Fig.~\ref{fig:1} illustrates the general idea of our distributed framework. Suppose that each institution maintains the ADNI genome-wide data for a few subjects. We first apply the distributed Lasso screening rule  to pre-identify inactive features and remove them from the training phase. Next, we employ the LQM on the reduced data matrices to perform collaborative analyses across different institutions. Finally, each institution obtains the learnt model and performs stability selection to rank the SNPs that may collectively affect AD. The process of stability selection is to count the frequency of nonzero entries in the solution vectors and select the most frequent ones as the top risk genes for AD. The whole learning procedure results in the same model for all institutions, and preserves data privacy at each of them.

\subsection{Local Query Model}
We apply a proximal gradient descent algorithm---the Iterative Shrinkage/Thresho-lding Algorithm (ISTA) \cite{daubechies2004iterative}---to solve problem (\ref{eq:1}). We define $g(x;A,y)=||Ax-y||^2_2$ as the least square loss function. The general updating rule of ISTA is:
\begin{equation}
x^{k+1} = \Gamma_{\lambda t_k}(x^k-t_k \nabla g(x^k;A,y)),
\label{eq:2}
\end{equation}
where $k$ is the iteration number, $t_k$ is an appropriate step size, and $\Gamma$ is the soft thresholding operator \cite{shalev2011stochastic} defined by $\Gamma_\alpha (x) = sign(x)\cdot (|x|-\alpha)_{+}$.

In view of (\ref{eq:2}), to solve (\ref{eq:1}), we need to compute the gradient of the loss function $\nabla g$, which equals to $A^T (Ax-y)$. However, because the data set $(A,y)$ is distributed to different institutions, we cannot compute the gradient directly. To address this challenge, we propose a Local Query Model to learn the model $x$ across multiple institutions without compromising data privacy.

In our study, each institution maintains its own data set $(A_i,y_i)$ to preserve their privacy. To avoid collecting all data matrices $A_i,i=1,2,...,m$ together, we can rewrite the problem (\ref{eq:1}) as the following equivalent formulation: $\min_{x} \sum_i^m g_i(x; A_i,y_i)+\lambda||x||_1: i=1,2,...,m,$ where $g_i(x; A_i, y_i)=\frac{1}{2} ||A_i x - y_i||_2^2$ is the least squares loss.

The key of LQM lies in the following decomposition: $\nabla g = A^T(Ax-y)=\sum_{i=1}^m A^T_i (A_ix-y_i)= \sum_{i=1}^m \nabla g_i$.
We use ``local institution'' to denote all the institutions and  ``global center'' to represent the place where intermediate results are calculated. The $i$th local institution computes $\nabla g_i = A^T_i (A_ix-y_i)$. Then, each local institution sends the partial gradient of the loss function to the global center. After gathering all the gradient information, the global center can compute the accurate gradient with respect to x by adding all $\nabla g_i$ together and send the updated gradient $\nabla g$ back to all the local institutions to compute $x$.

The master (global center) only servers as the computation center and does not store any data sets. Although the master gets $g_i$,  it could not reconstruct $A_i$ and $y_i$. Let $g_i^k$ denote the $k$th iteration of $g_i$. Suppose $x$ is initialized to be zero, $g_i^1=-A_i^T y_i$ and $g_i^k= A_i(A_i^T x^k-y_i)$.  We get $A^T_i A_i x $ by $g_i^k-g_i^1$ but $A_i$ can not be reconstructed since updating and storing $x$ only happens in the workers (local institution).  As a result, LQM can properly maintain data privacy for all the institutions.

\subsection{Safe Screening Rules for Lasso}

The dual problem of Lasso (\ref{eq:1}) can be formulated as the following equation:
\begin{equation}
\sup_{\theta}\left\{\frac{1}{2}||y||_2^2-\frac{\lambda^2}{2}||\theta-\frac{y}{\lambda}||_2^2 : \left|[A]_j^T \theta\right|\leq 1, j = 1,2,...,p \right\},
\label{eq:6}
\end{equation}
where $\theta$ is the dual variable and $[A]_j$ denotes the $j$th column of $A$. Let $\theta^*(\lambda)$ be the optimal solution of problem (\ref{eq:6}) and $x^*(\lambda)$ denotes the optimal solution of problem (\ref{eq:1}). The Karush--Kuhn--Tucker (KKT) conditions are given by:
\begin{equation}
y = Ax^*(\lambda)+\lambda \theta^*(\lambda),
\label{eq:7}
\end{equation}
\begin{equation}
 [A]_j^T \theta^*(\lambda) \in \left\{
\begin{array}{l}
\text{sign} ([x^*(\lambda)]_j), \quad \text{If }[x^*(\lambda)]_j \neq 0,
\\
{[-1,1]}, \quad \quad \quad \quad \text{If } [x^*(\lambda)]_j=0,
\end{array}
\right.
\label{eq:8}
\end{equation}
where $[x^*(\lambda)]_k$ denotes the $k$th component of $x^*(\lambda)$. In view of the KKT condition in equation (\ref{eq:8}), the following rule holds: $\left| [A]_j^T \theta^*(\lambda) \right|<1 \Rightarrow [x^*(\lambda)]_j=0 \Rightarrow x_j \text{ is an inactive feature}$.

The inactive features have zero components in the optimal solution vector $x^*(\lambda)$ so that we can remove them from the optimization without sacrificing the accuracy of the optimal value in the objective function (\ref{eq:1}). We call this kind of screening methods as \textit{Safe Screening Rules}. SAFE \cite{ghaoui2010safe} is one of highly efficient safe screening methods. In SAFE, the $j$th entry of $x^*(\lambda)$ is discarded when
\begin{equation}
\left|[A]_j^Ty\right|<\lambda - ||[A]_j||_2||y||_2\frac{\lambda_{\max}-\lambda}{\lambda_{\max}},
\end{equation}
where $\lambda_{\max} = \max_j \left| [A]_j^T y \right|$. As a result, the optimization can be performed on the reduced data matrix $\widetilde{A}$ and the original problem (\ref{eq:1}) can be reformulated as:
\begin{equation}
 \min\limits_{\widetilde{x}} {\frac{1}{2}||\widetilde{A}\widetilde{x}-y||_2^2+ \lambda||\widetilde{x}||_1:  \widetilde{x}\in  \mathbb{R}^{\widetilde{p}} } \text{ and }\widetilde{A}\in \mathbb{R}^{n\times \widetilde{p}},
\end{equation}
where $\widetilde{p}$ is the number of remaining features after employing safe screening rules. The optimization is performed on a reduced feature matrix, accelerating the whole learning process significantly.

\subsection{Distributed Safe Screening Rules for Lasso}
As data are distributed to different institutions, we develop a family of distributed Lasso screening rule to identify and discard inactive features in a distributed environment. Suppose $i$th institution holds the data set $(A_i, y_i)$, we summarize a distributed version of SAFE screening rules (D-SAFE) as follows:

Step 1:  $ Q_i=[A_i]^T y_i$, update $Q = \sum_i^m Q_i$ by LQM.

Step 2:  $ \lambda_{\max} = \max_j \left| [Q]_j  \right|.$

Step 3:   If $\left|[A]_j^Ty\right|<\lambda - ||[A]_j||_2||y||_2\frac{\lambda_{\max}-\lambda}{\lambda_{\max}}$, discard $j$th feature.

To compute $||[A]_j||_2$ in Step $3$, we first compute $H_i=||[A_i]_j||_2^2$ and perform LQM to compute $H$ by $H=\sum_i^m H_i$. Then, we have $||[A_i]_j||_2=\sqrt H$. Similarly, we can compute $||y||_2$ in Step $3$. As the data communication only requires intermediate results, D-SAFE preserves the data privacy at each institution.

To tune the value of $\lambda$, commonly used methods such as cross validation need to solve the Lasso problem along a sequence of parameters $\lambda_0> \lambda_1>...>\lambda_{\kappa}$, which can be very time-consuming. Enhanced Dual Polytope Projection (EDPP) \cite{wang2013lasso} is a highly efficient safe screening rules. Implementation details of EDPP is available on the GitHub: http://dpc-screening.github.io/lasso.html.

To address the problem of data privacy,  we propose a distributed Lasso screening rule, termed Distributed Enhanced Dual Polytope Projection (D-EDPP), to identify and discard inactive features along a sequence of parameter values in a distributed manner. The idea of D-EDPP is similar to LQM. Specifically, to update the global variables, we apply LQM to query each local center for intermediate results--computed locally--and we aggregate them at global center. After obtaining the reduced matrix for each institution, we apply LQM to solve the Lasso problem on the reduced data set $\widetilde{A}_i$, $i=1,...,m$. We assume that $j$ indicates the $j$th column in $A$, $j=1,...,p$, where $p$ is the number of features. We summarize the proposed D-EDPP in Algorithm 1.

To calculate $R$, we apply LQM through aggregating all the $R_i$ together in the global center by $R=\sum_i^m R_i$ and send $R$ back to every institution. The same approach is used to calculate $T$, $S$ and $w$ in D-EDPP. The calculation of $||[A]_j||_2$ and $||v^{\bot}_2(\lambda_{k},\lambda_{k-1})||_2$ follows the same way in D-SAFE. The discarding result of $\lambda_k$ relies on the previous optimal solution $x^*(\lambda_{k-1})$. Especially, $\lambda_k$ equals to $\lambda_{\max}$ when $k$ is zero. Thus, we identify all the elements to be zero at $x^*(\lambda_0)$. When $k$ is $1$, we can perform screening based on $x^*(\lambda_0)$.
\begin{algorithm}[t]
\caption{Distributed Enhanced Dual Polytope Projection (D-EDPP)}
\begin{algorithmic}[1]
   \Require A set of data pairs $\{ (A_1, y_1), (A_2, y_2),..., (A_n, y_n) \}$ and $i$th institution holds the data pair $(A_i, y_i)$. A sequence of parameters: $\lambda_{\max} = \lambda_0> \lambda_1>...>\lambda_{\kappa}$.
   \Ensure The learnt models: $\{x^*(\lambda_0), x^*(\lambda_1),..., x^*(\lambda_{\kappa})\}$.
   \State Perform the computation on $n$ institutions. For the $i$th institution:
   \State \quad Let $R_i = A^T_i y_i$, compute $R=\sum_i^m R_i$ by LQM.  Then we get $\lambda_{\max}$ by $||R||_{\infty}$.
   \State \quad $J = \arg\max_j |R|$, $v_i=[A_i]_J$ where $[A_i]_J$ is the $J$th column of $A_i$.
   \State \quad Let $\lambda_{0}\in (0, \lambda_{\max}]$ and $\lambda \in (0, \lambda_{0}]$.
   \State \quad  \quad $\theta_i(\lambda) = \left\{
			\begin{array}{l}
			\frac{y_i}{\lambda_{\max}}, \quad \quad  \text{    if }\lambda=\lambda_{\max}, \\
			\frac{y_i-A_i x^*(\lambda)}{\lambda}, \quad \text{if }\lambda \in (0,\lambda_{\max}),
			\end{array}
			\right.$
   \State \quad \quad $T_i = v^T_i * y_i$, compute $T=\sum_i^m T_i$ by LQM.
   \State \quad  \quad $v_1(\lambda_0)_i = \left\{
			\begin{array}{l}
			\text{sign}(T)*v_i, \quad  \text{if }\lambda_0=\lambda_{\max}, \\
			\frac{y_i}{\lambda_0} - \theta_i(\lambda_0), \quad \text{if }\lambda_0 \in (0,\lambda_{\max}),
			\end{array}
			\right.$
   \State \quad \quad $v_2(\lambda, \lambda_0)_i = \frac{y_i}{\lambda} - \theta_i (\lambda_0)$, $S_i = ||v_1(\lambda_0)_i||_2^2$, compute $S=\sum_i^m S_i$ by LQM.
   \State \quad \quad $v_2^{\bot}(\lambda, \lambda_0)_i = v_2(\lambda, \lambda_0)_i - \frac{<v_1(\lambda_0)_i, v_2(\lambda, \lambda_0)_i>}{S} v_1(\lambda_0)_i$.
   \smallskip
   \State \quad Given a sequence of parameters: $\lambda_{\max} = \lambda_0> \lambda_1>...>\lambda_{\kappa}$, for $k\in [1, \kappa]$, we make a prediction of screening on $\lambda_k$ if $x^*(\lambda_{k-1})$ is known:
    \smallskip
   \State \quad \textbf{for} j=1 \textbf{to} p \textbf{do}
   \State \quad \quad $w_i=[A_i]^T_j(\theta_i(\lambda_{k-1})+\frac{1}{2} v_2^{\bot}(\lambda_k, \lambda_{k-1})_i)$, compute $w=\sum_i^m w_i$ by LQM.
   \State \quad \quad \textbf{if}  $w < 1-\frac{1}{2}||v^{\bot}_2(\lambda_{k},\lambda_{k-1})||_2||[A]_j||_2$ \textbf{then}
         \smallskip
   \State \quad \quad \quad We identify $[x^*(\lambda_{k})]_j = 0$.
    \State \quad  \textbf{end for}
\end{algorithmic}
\label{alg:2}
\end{algorithm}

\subsection{Local Query Model for Lasso}
To further accelerate the learning process, we apply FISTA \cite{beck2009fast} to solve the Lasso problem in a distributed manner. The convergence rate of FISTA is $O(1/k^2)$ compared to $O(1/k)$ of ISTA, where $k$ is the iteration number. We integrate FISTA with LQM (F-LQM) to solve the Lasso problem on the reduced matrix $\widetilde{A}_i$. We summarize the updating rule of F-LQM in $k$th iteration as follows:

Step 1:  $\nabla g_i^k = \widetilde{A}_i^T ( \widetilde{A}_i  x^k-y_i )$, update $\nabla g^k = \sum_i^m \nabla g_i^k$ by LQM.

Step 2:  $z^k = \Gamma_{\lambda t_k} (x^k - t_k \nabla g^k)$ and $t_{k+1}=\frac{1+\sqrt {1+4t^2_k} }{2}$.

Step 3:  $x^{k+1} = z^k + \frac{t_k-1}{t_{k+1}} (z^k-z^{k-1})$.

The matrix $\widetilde{A}_i$ denotes the reduced matrix for the $i$th institution obtained by D-EDPP rule. We repeat this procedure until a satisfactory global model is obtained. Step 1 calculates $ \nabla g_i^k $ from local data $( \widetilde{A}_i, y_i)$. Then, each institution performs LQM to get the gradient $\nabla g^k$ based on (5). Step 2 updates the auxiliary variables $z^k$ and step size $t_k$. Step 3 updates the model $x$. Similar to LQM, the data privacy of institutions are well preserved by F-LQM.

\section{Experiment}

 We implement the proposed framework across three institutions on a state-of-the-art distributed platform---Apache Spark---a fast and efficient distributed platform for large-scale data computing. Experiment shows the efficiency and effectiveness of proposed models.

\begin{figure}[t]
\centering
\includegraphics[height=5cm]{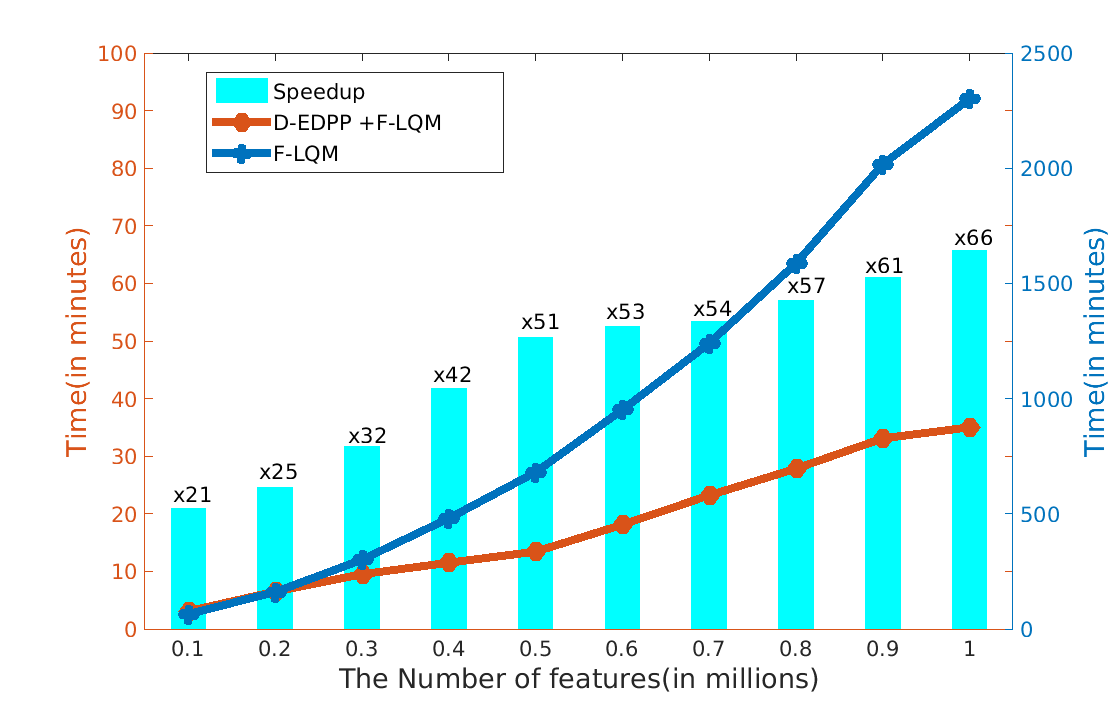}
\caption{Running time comparison of Lasso with and without D-EDPP rules.}
\label{fig:2}
\end{figure}

\subsection{Comparison of Lasso with and without D-EDPP rule}

We choose the volume of lateral ventricle as variables being predicted in trials containing 717 subjects by removing subjects without labels. The volumes of brain regions were extracted from each subject's T1 MRI scan using Freesurfer: http://freesurfer.net. We evaluate the efficiency of D-EDPP across three research institutions that maintain 326, 215, and 176 subjects, respectively. The subjects are stored as HDFS files. We solve the Lasso problem along a sequence of 100 parameter values equally spaced on the linear scale of $\lambda/\lambda_{\max}$  from 1.00 to 0.05. We randomly select 0.1 million to 1 million features by applying F-LQM since \cite{beck2009fast} proved that FISTA converges faster than ISTA. We report the result in Fig. \ref{fig:2} and achieved about a speedup of 66-fold compared to F-LQM.

\subsection{Stability selection for top risk genetic factors}

We employ stability selection \cite{meinshausen2010stability,yang2015detecting} with D-EDPP+F-LQM to select top risk SNPs from the entire GWAS with 5,906,152 features. We conduct four groups of trials in Table \ref{tab:1}. In each trial, D-EDPP+F-LQM is carried out along a 100 linear-scale sequence from 1 to 0.05. We simulate this 200 times and perform on 500 of subjects in each round. Table \ref{tab:1} shows the top 5 selected SNPs. APOE, one of the top genetic risk factors for AD \cite{liu2013apolipoprotein}, is ranked \#1 for three groups.

\begin{table}[t]
\centering
\caption{Top 5 selected risk SNPs associated with diagnose, the volume of hippocampal, entorhinal cortex, and lateral ventricle at baseline, based on ADNI. }
\label{tab:1}
\centering
\begin{tabular}{|C{0.5cm}|C{0.6cm}|C{1.5cm}|C{1.58cm}|C{1.58cm}|C{0.5cm}|C{0.6cm}|C{1.5cm}|C{1.58cm}|C{1.55cm}|} \hline
       \multicolumn{5}{|c|}{Diagnose at baseline}& \multicolumn{5}{c|}{Hippocampus at baseline} \\ \hline
  No. & Chr & Position  & RS\_ID  & Gene &No.  & Chr & Position  & RS\_ID & Gene\\ \hline
  1 & 19 & 45411941 & rs429358 & APOE & 1 & 19 & 45411941 & rs429358 & APOE\\ \hline
  2 & 19 & 45410002 & rs769449 & APOE & 2 & 8  & 145158607 & rs34173062 & SHARPIN\\ \hline
  3 & 12 & 9911736 & rs3136564 & CD69 & 3 & 11 & 11317240 & rs10831576 & GALNT18\\ \hline
  4 & 1 & 172879023 & rs2227203 & unknown & 4 & 10 & 71969989 & rs12412466 & PPA1 \\ \hline
  5 & 20 & 58267891 & rs6100558 & PHACTR3 & 5 & 6 & 168107162 & rs71573413 & unknown \\ \hline
\end{tabular}
\begin{tabular}{|C{0.5cm}|C{0.6cm}|C{1.5cm}|C{1.58cm}|C{1.58cm}|C{0.5cm}|C{0.6cm}|C{1.5cm}|C{1.58cm}|C{1.55cm}|} \hline
       \multicolumn{5}{|c|}{Entorhinal cortex at baseline}& \multicolumn{5}{c|}{Lateral ventricle at baseline} \\ \hline
  No. & Chr & Position  & RS\_ID  & Gene &No. & Chr & Position  & RS\_ID & Gene \\ \hline
  1 & 19 & 45411941 & rs429358 & APOE & 1 & Y & 3164319 & rs2261174 & unknown \\ \hline
  2 & 15  & 89688115 & rs8025377 & ABHD2 & 2 & 10 & 62162053 & rs10994327 & ANK3 \\ \hline
  3 & Y & 10070927 & rs79584829 & unknown & 3 & Y & 13395084 & rs62610496 & unknown \\ \hline
  4 & 14 & 47506875 & rs41354245 & MDGA2 & 4 & 1 & 77895410 & rs2647521 & AK5 \\ \hline
  5 & 3 & 30106956 & rs55904134 & unknown & 5 & 1 & 114663751 & rs2629810 & SYT6 \\ \hline
\end{tabular}
\end{table}



\subsubsection{Acknowledgments}
This work was supported in part by NIH Big Data to Knowledge (BD2K) Center of Excellence grant U54 EB020403,  funded  by  a  cross-NIH  consortium  including NIBIB and NCI.

\bibliographystyle{splncs03}
\bibliography{miccai}
\end{document}